\documentclass[11pt,a4paper]{article}
\pdfoutput=1
\usepackage[hyperref]{naaclhlt2019}
\usepackage{times}
\usepackage{latexsym}

\usepackage{url}

\aclfinalcopy 
\def\aclpaperid{1506} 

\usepackage{graphicx}
\usepackage{amsmath}
\usepackage{amssymb}
\usepackage{amsmath,amsfonts,bm}

\def\modelname{\textsc{AutoSeM}}

\usepackage{color}

\usepackage{algorithm}
\usepackage{algorithmicx}
\usepackage{algpseudocode}

\title{\modelname: Automatic Task Selection and Mixing in Multi-Task Learning}

\author{Han Guo \;\;\;\;\;\;\; Ramakanth Pasunuru \;\;\;\;\;\;\; Mohit Bansal \\
  UNC Chapel Hill \\
  {\tt \{hanguo, ram, mbansal\}@cs.unc.edu} \\
 }

\date{}

\begin{document}
\maketitle
\begin{abstract}
  Multi-task learning (MTL) has achieved success over a wide range of problems, where the goal is to improve the performance of a primary task using a set of relevant auxiliary tasks. However, when the usefulness of the auxiliary tasks w.r.t. the primary task is not known a priori, the success of MTL models depends on the correct choice of these auxiliary tasks and also a balanced mixing ratio of these tasks during alternate training. These two problems could be resolved via manual intuition or hyper-parameter tuning over all combinatorial task choices, but this introduces inductive bias or is not scalable when the number of candidate auxiliary tasks is very large.  To address these issues, we present \modelname{}, a two-stage MTL pipeline, where the first stage automatically selects the most useful auxiliary tasks via a Beta-Bernoulli multi-armed bandit with Thompson Sampling, and the second stage learns the training mixing ratio of these selected auxiliary tasks via a Gaussian Process based Bayesian optimization framework. We conduct several MTL experiments on the GLUE language understanding tasks, and show that our \modelname{} framework can successfully find relevant auxiliary tasks and automatically learn their mixing ratio, achieving significant performance boosts on several primary tasks. Finally, we present ablations for each stage of \modelname{} and analyze the learned auxiliary task choices.
\end{abstract}

\section{Introduction}
Multi-task Learning (MTL)~\cite{caruana1997multitask} is an inductive transfer mechanism which leverages information from related tasks to improve the primary model's generalization performance. It achieves this goal by training multiple tasks in parallel while sharing representations, where the training signals from the auxiliary tasks can help improve the performance of the primary task. Multi-task learning has been applied to a wide range of natural language processing problems~\cite{luong2015multi,pasunuru2017multitask,hashimoto2017ajm,ruder2017sluice,kaiser2017one,mccann2018natural}.
Despite its impressive performance, the design of a multi-task learning system is non-trivial.
In the context of improving the primary task's performance using knowledge from other auxiliary tasks~\cite{luong2015multi,pasunuru2017multitask}, two major challenges include selecting the most relevant auxiliary tasks and also learning the balanced mixing ratio for synergized training of these tasks. 
One can achieve this via manual intuition or hyper-parameter tuning over all combinatorial task choices, but this introduces human inductive bias or is not scalable when the number of candidate auxiliary tasks is considerable. To this end, we present \modelname{}, a two-stage Bayesian optimization pipeline to this problem. 

In our \modelname{} framework\footnote{We make all our code and models publicly available at: \url{https://github.com/HanGuo97/AutoSeM}}, the first stage addresses automatic task selection from a pool of auxiliary tasks. For this, we use a non-stationary multi-armed bandit controller (MAB)~\cite{bubeck2012regret,raj2017taming} that dynamically alternates among task choices within the training loop, and eventually returns estimates of the utility of each task w.r.t. the primary task. We model the utility of each task as a Beta distribution, whose expected value can be interpreted as the probability of each task making a non-negative contribution to the training performance of the primary task. Further, we model the observations as Bernoulli variables so that the posterior distribution is also Beta-distributed. We use Thompson sampling~\cite{chapelle2011empirical,russo2018tutorial} to trade off exploitation and exploration.

The second stage then takes the auxiliary tasks selected in the first stage and automatically learns the training mixing ratio of these tasks, through the framework of Bayesian optimization, by modeling the performance of each mixing ratio as a sample from a Gaussian Process (GP) to sequentially search for the optimal values~\cite{rasmussen2004gaussian,snoek2012practical}.
For the covariance function in the GP, we use the Matern kernel which is parameterized by a smoothness hyperparameter so as to control the level of differentiability of the samples from GP. 
Further, following~\citet{hoffman2011portfolio}, we use a portfolio of optimistic and improvement-based policies as acquisition functions~\cite{shahriari2016taking} for selecting the next sample point from the GP search space.

We conduct several experiments on the GLUE natural language understanding benchmark~\cite{wang2018glue}, where we choose each of RTE, MRPC, QNLI, CoLA, and SST-2 as the primary task, and treat the rest of the classification tasks from the GLUE benchmark as candidate auxiliary tasks. Results show that our \modelname{} framework can successfully find useful auxiliary tasks and automatically learn their mixing ratio, achieving significant performance boosts on top of strong baselines for several primary tasks, e.g., 5.2\% improvement on QNLI, 4.7\% improvement on RTE, and 2.8\%/0.8\% improvement on MRPC.

We also ablate the usefulness of our two stages of auxiliary task selection and automatic mixing ratio learning. The first ablation removes the task selection stage and instead directly performs the second GP mixing ratio learning stage on all auxiliary tasks. The second ablation performs the task selection stage (with multi-armed bandit) but replaces the second stage Gaussian Process with manual tuning on the selected tasks. Our 2-stage model performs better than both these ablations, showing that both of our stages are crucial. Further, we also discuss the learned auxiliary task choices in terms of their intuitive relevance w.r.t. the corresponding primary task.

\section{Related Work}
\label{section:related-works}

Multi-task learning~\cite{caruana1998multitask}, known for improving the generalization performance of a task with auxiliary tasks, has successfully been applied to many domains of machine learning, including natural language processing~\cite{collobert2008unified,girshick2015fast,luong2015multi,pasunuru2017multitask,Pasunuru2017TowardsIA}, computer vision~\cite{misra2016cross,kendall2017multi,dai2016instance}, and reinforcement learning~\cite{teh2017distral,parisotto2015actor,jaderberg2016reinforcement}. Although there are many variants of multi-task learning~\cite{ruder2017sluice,hashimoto2017ajm,luong2015multi,mccann2018natural}, our goal is to improve the performance of a primary task using a set of relevant auxiliary tasks, where different tasks share some common model parameters with alternating mini-batches optimization, similar to~\citet{luong2015multi}. 

To address the problem of automatic shared parameter selection,~\newcite{ruder2017learning} automatically learned the latent multi-task sharing architecture, and~\newcite{xiao2018gated} used a gate mechanism that filters the feature flows between tasks.
On the problem of identifying task relatedness,~\citet{ben2003exploiting} provided a formal framework for task relatedness and derived generalization error bounds for learning of multiple tasks.
~\citet{bingel2017identifying} explored task relatedness via exhaustively experimenting with all possible two task tuples in a non-automated multi-task setup. Other related works explored data selection, where the goal is to select or reorder the examples from one or more domains (usually in a single task) to either improve the training efficiency or enable better transfer learning. These approaches have been applied in machine translation~\cite{van2017dynamic}, language models~\cite{moore2010intelligent,duh2013adaptation}, dependency parsing~\cite{sogaard2011data}, etc. In particular,~\citet{ruder2017learning2} used Bayesian optimization to select relevant training instances for transfer learning, and~\citet{tsvetkov2016learning} applied it to learn a curriculum for training word embeddings via reordering data.~\citet{graves2017automated} used the bandit approach (Exp3.S algorithm) in the context of automated curriculum learning, but in our work, we have two stages with each stage addressing a different problem (automatic task selection and learning of the training mixing ratio). Recently,~\newcite{sharma2017online} used multi-armed bandits (MAB) to learn the choice of hard vs. easy domain data selection as input feed for the model.
\newcite{guo2018dynamic} used MAB to effectively switch across tasks in a dynamic multi-task learning setup.
In our work, we use MAB with Thompson Sampling for the novel paradigm of automatic auxiliary task selection; and next, we use a Matern-kernel Gaussian Process to automatically learn an exact (static) mixing ratio (i.e., relatedness ratio) for the small number of selected tasks.

Many control problems can be cast as a multi-armed bandits problem, where the goal of the agent is to select the arm/action from one of the $N$ choices that minimizes the regrets~\cite{bubeck2012regret}. One problem in bandits learning is the trade-off between exploration and exploitation, where the agent needs to make a decision between taking the action that yields the best payoff on current estimates or exploring new actions whose payoffs are not yet certain.
Many previous works have explored various exploration and exploitation strategies to minimize regret, including Boltzmann exploration~\cite{kaelbling1996reinforcement}, adversarial bandits~\cite{auer2002nonstochastic}, UCB~\cite{auer2002finite}, and information gain using variational approaches~\cite{houthooft2016vime}. In this work, for task selection, we use Thompson Sampling~\citep{russo2018tutorial,chapelle2011empirical}, an algorithm for sequential decision making problems, which addresses a broad range of problems in a computationally efficient manner and is therefore enjoying wide use.

Gaussian Process (GP) is a non-parametric Bayesian approach, and it can capture a wide variety of underlying functions or relations between inputs and outputs by taking advantage of the full information provided by the history of observations and is thus very data-efficient~\cite{rasmussen2004gaussian,shahriari2016taking,schulz2018tutorial}. Gaussian Processes have been widely used as a black-box optimizer and hyper-parameter optimization~\cite{snoek2012practical,brochu2010tutorial,GPflowOpt2017,cully2018limbo,swersky2013multi,golovin2017google}. In our work, we use Gaussian Process for automatic learning of the multi-task mixing ratio in our stage-2 among the selected tasks from stage-1.



\newcommand{\figleft}{{\em (Left)}}
\newcommand{\figcenter}{{\em (Center)}}
\newcommand{\figright}{{\em (Right)}}
\newcommand{\figtop}{{\em (Top)}}
\newcommand{\figbottom}{{\em (Bottom)}}
\newcommand{\captiona}{{\em (a)}}
\newcommand{\captionb}{{\em (b)}}
\newcommand{\captionc}{{\em (c)}}
\newcommand{\captiond}{{\em (d)}}

\newcommand{\newterm}[1]{{\bf #1}}

\def\figref#1{figure~\ref{#1}}
\def\Figref#1{Figure~\ref{#1}}
\def\twofigref#1#2{figures \ref{#1} and \ref{#2}}
\def\quadfigref#1#2#3#4{figures \ref{#1}, \ref{#2}, \ref{#3} and \ref{#4}}
\def\secref#1{section~\ref{#1}}
\def\Secref#1{Section~\ref{#1}}
\def\twosecrefs#1#2{sections \ref{#1} and \ref{#2}}
\def\secrefs#1#2#3{sections \ref{#1}, \ref{#2} and \ref{#3}}
\def\eqref#1{equation~\ref{#1}}
\def\Eqref#1{Equation~\ref{#1}}
\def\plaineqref#1{\ref{#1}}
\def\chapref#1{chapter~\ref{#1}}
\def\Chapref#1{Chapter~\ref{#1}}
\def\rangechapref#1#2{chapters\ref{#1}--\ref{#2}}
\def\algref#1{algorithm~\ref{#1}}
\def\Algref#1{Algorithm~\ref{#1}}
\def\twoalgref#1#2{algorithms \ref{#1} and \ref{#2}}
\def\Twoalgref#1#2{Algorithms \ref{#1} and \ref{#2}}
\def\partref#1{part~\ref{#1}}
\def\Partref#1{Part~\ref{#1}}
\def\twopartref#1#2{parts \ref{#1} and \ref{#2}}

\def\ceil#1{\lceil #1 \rceil}
\def\floor#1{\lfloor #1 \rfloor}
\def\1{\bm{1}}
\newcommand{\train}{\mathcal{D}}
\newcommand{\valid}{\mathcal{D_{\mathrm{valid}}}}
\newcommand{\test}{\mathcal{D_{\mathrm{test}}}}

\def\eps{{\epsilon}}

\def\reta{{\textnormal{$\eta$}}}
\def\ra{{\textnormal{a}}}
\def\rb{{\textnormal{b}}}
\def\rc{{\textnormal{c}}}
\def\rd{{\textnormal{d}}}
\def\re{{\textnormal{e}}}
\def\rf{{\textnormal{f}}}
\def\rg{{\textnormal{g}}}
\def\rh{{\textnormal{h}}}
\def\ri{{\textnormal{i}}}
\def\rj{{\textnormal{j}}}
\def\rk{{\textnormal{k}}}
\def\rl{{\textnormal{l}}}
\def\rn{{\textnormal{n}}}
\def\ro{{\textnormal{o}}}
\def\rp{{\textnormal{p}}}
\def\rq{{\textnormal{q}}}
\def\rr{{\textnormal{r}}}
\def\rs{{\textnormal{s}}}
\def\rt{{\textnormal{t}}}
\def\ru{{\textnormal{u}}}
\def\rv{{\textnormal{v}}}
\def\rw{{\textnormal{w}}}
\def\rx{{\textnormal{x}}}
\def\ry{{\textnormal{y}}}
\def\rz{{\textnormal{z}}}

\def\rvepsilon{{\mathbf{\epsilon}}}
\def\rvtheta{{\mathbf{\theta}}}
\def\rva{{\mathbf{a}}}
\def\rvb{{\mathbf{b}}}
\def\rvc{{\mathbf{c}}}
\def\rvd{{\mathbf{d}}}
\def\rve{{\mathbf{e}}}
\def\rvf{{\mathbf{f}}}
\def\rvg{{\mathbf{g}}}
\def\rvh{{\mathbf{h}}}
\def\rvu{{\mathbf{i}}}
\def\rvj{{\mathbf{j}}}
\def\rvk{{\mathbf{k}}}
\def\rvl{{\mathbf{l}}}
\def\rvm{{\mathbf{m}}}
\def\rvn{{\mathbf{n}}}
\def\rvo{{\mathbf{o}}}
\def\rvp{{\mathbf{p}}}
\def\rvq{{\mathbf{q}}}
\def\rvr{{\mathbf{r}}}
\def\rvs{{\mathbf{s}}}
\def\rvt{{\mathbf{t}}}
\def\rvu{{\mathbf{u}}}
\def\rvv{{\mathbf{v}}}
\def\rvw{{\mathbf{w}}}
\def\rvx{{\mathbf{x}}}
\def\rvy{{\mathbf{y}}}
\def\rvz{{\mathbf{z}}}

\def\erva{{\textnormal{a}}}
\def\ervb{{\textnormal{b}}}
\def\ervc{{\textnormal{c}}}
\def\ervd{{\textnormal{d}}}
\def\erve{{\textnormal{e}}}
\def\ervf{{\textnormal{f}}}
\def\ervg{{\textnormal{g}}}
\def\ervh{{\textnormal{h}}}
\def\ervi{{\textnormal{i}}}
\def\ervj{{\textnormal{j}}}
\def\ervk{{\textnormal{k}}}
\def\ervl{{\textnormal{l}}}
\def\ervm{{\textnormal{m}}}
\def\ervn{{\textnormal{n}}}
\def\ervo{{\textnormal{o}}}
\def\ervp{{\textnormal{p}}}
\def\ervq{{\textnormal{q}}}
\def\ervr{{\textnormal{r}}}
\def\ervs{{\textnormal{s}}}
\def\ervt{{\textnormal{t}}}
\def\ervu{{\textnormal{u}}}
\def\ervv{{\textnormal{v}}}
\def\ervw{{\textnormal{w}}}
\def\ervx{{\textnormal{x}}}
\def\ervy{{\textnormal{y}}}
\def\ervz{{\textnormal{z}}}

\def\rmA{{\mathbf{A}}}
\def\rmB{{\mathbf{B}}}
\def\rmC{{\mathbf{C}}}
\def\rmD{{\mathbf{D}}}
\def\rmE{{\mathbf{E}}}
\def\rmF{{\mathbf{F}}}
\def\rmG{{\mathbf{G}}}
\def\rmH{{\mathbf{H}}}
\def\rmI{{\mathbf{I}}}
\def\rmJ{{\mathbf{J}}}
\def\rmK{{\mathbf{K}}}
\def\rmL{{\mathbf{L}}}
\def\rmM{{\mathbf{M}}}
\def\rmN{{\mathbf{N}}}
\def\rmO{{\mathbf{O}}}
\def\rmP{{\mathbf{P}}}
\def\rmQ{{\mathbf{Q}}}
\def\rmR{{\mathbf{R}}}
\def\rmS{{\mathbf{S}}}
\def\rmT{{\mathbf{T}}}
\def\rmU{{\mathbf{U}}}
\def\rmV{{\mathbf{V}}}
\def\rmW{{\mathbf{W}}}
\def\rmX{{\mathbf{X}}}
\def\rmY{{\mathbf{Y}}}
\def\rmZ{{\mathbf{Z}}}

\def\ermA{{\textnormal{A}}}
\def\ermB{{\textnormal{B}}}
\def\ermC{{\textnormal{C}}}
\def\ermD{{\textnormal{D}}}
\def\ermE{{\textnormal{E}}}
\def\ermF{{\textnormal{F}}}
\def\ermG{{\textnormal{G}}}
\def\ermH{{\textnormal{H}}}
\def\ermI{{\textnormal{I}}}
\def\ermJ{{\textnormal{J}}}
\def\ermK{{\textnormal{K}}}
\def\ermL{{\textnormal{L}}}
\def\ermM{{\textnormal{M}}}
\def\ermN{{\textnormal{N}}}
\def\ermO{{\textnormal{O}}}
\def\ermP{{\textnormal{P}}}
\def\ermQ{{\textnormal{Q}}}
\def\ermR{{\textnormal{R}}}
\def\ermS{{\textnormal{S}}}
\def\ermT{{\textnormal{T}}}
\def\ermU{{\textnormal{U}}}
\def\ermV{{\textnormal{V}}}
\def\ermW{{\textnormal{W}}}
\def\ermX{{\textnormal{X}}}
\def\ermY{{\textnormal{Y}}}
\def\ermZ{{\textnormal{Z}}}

\def\vzero{{\bm{0}}}
\def\vone{{\bm{1}}}
\def\vmu{{\bm{\mu}}}
\def\vtheta{{\bm{\theta}}}
\def\va{{\bm{a}}}
\def\vb{{\bm{b}}}
\def\vc{{\bm{c}}}
\def\vd{{\bm{d}}}
\def\ve{{\bm{e}}}
\def\vf{{\bm{f}}}
\def\vg{{\bm{g}}}
\def\vh{{\bm{h}}}
\def\vi{{\bm{i}}}
\def\vj{{\bm{j}}}
\def\vk{{\bm{k}}}
\def\vl{{\bm{l}}}
\def\vm{{\bm{m}}}
\def\vn{{\bm{n}}}
\def\vo{{\bm{o}}}
\def\vp{{\bm{p}}}
\def\vq{{\bm{q}}}
\def\vr{{\bm{r}}}
\def\vs{{\bm{s}}}
\def\vt{{\bm{t}}}
\def\vu{{\bm{u}}}
\def\vv{{\bm{v}}}
\def\vw{{\bm{w}}}
\def\vx{{\bm{x}}}
\def\vy{{\bm{y}}}
\def\vz{{\bm{z}}}

\def\evalpha{{\alpha}}
\def\evbeta{{\beta}}
\def\evepsilon{{\epsilon}}
\def\evlambda{{\lambda}}
\def\evomega{{\omega}}
\def\evmu{{\mu}}
\def\evpsi{{\psi}}
\def\evsigma{{\sigma}}
\def\evtheta{{\theta}}
\def\eva{{a}}
\def\evb{{b}}
\def\evc{{c}}
\def\evd{{d}}
\def\eve{{e}}
\def\evf{{f}}
\def\evg{{g}}
\def\evh{{h}}
\def\evi{{i}}
\def\evj{{j}}
\def\evk{{k}}
\def\evl{{l}}
\def\evm{{m}}
\def\evn{{n}}
\def\evo{{o}}
\def\evp{{p}}
\def\evq{{q}}
\def\evr{{r}}
\def\evs{{s}}
\def\evt{{t}}
\def\evu{{u}}
\def\evv{{v}}
\def\evw{{w}}
\def\evx{{x}}
\def\evy{{y}}
\def\evz{{z}}

\def\mA{{\bm{A}}}
\def\mB{{\bm{B}}}
\def\mC{{\bm{C}}}
\def\mD{{\bm{D}}}
\def\mE{{\bm{E}}}
\def\mF{{\bm{F}}}
\def\mG{{\bm{G}}}
\def\mH{{\bm{H}}}
\def\mI{{\bm{I}}}
\def\mJ{{\bm{J}}}
\def\mK{{\bm{K}}}
\def\mL{{\bm{L}}}
\def\mM{{\bm{M}}}
\def\mN{{\bm{N}}}
\def\mO{{\bm{O}}}
\def\mP{{\bm{P}}}
\def\mQ{{\bm{Q}}}
\def\mR{{\bm{R}}}
\def\mS{{\bm{S}}}
\def\mT{{\bm{T}}}
\def\mU{{\bm{U}}}
\def\mV{{\bm{V}}}
\def\mW{{\bm{W}}}
\def\mX{{\bm{X}}}
\def\mY{{\bm{Y}}}
\def\mZ{{\bm{Z}}}
\def\mBeta{{\bm{\beta}}}
\def\mPhi{{\bm{\Phi}}}
\def\mLambda{{\bm{\Lambda}}}
\def\mSigma{{\bm{\Sigma}}}

\newcommand{\tens}[1]{\bm{\mathsfit{#1}}}
\def\tA{{\tens{A}}}
\def\tB{{\tens{B}}}
\def\tC{{\tens{C}}}
\def\tD{{\tens{D}}}
\def\tE{{\tens{E}}}
\def\tF{{\tens{F}}}
\def\tG{{\tens{G}}}
\def\tH{{\tens{H}}}
\def\tI{{\tens{I}}}
\def\tJ{{\tens{J}}}
\def\tK{{\tens{K}}}
\def\tL{{\tens{L}}}
\def\tM{{\tens{M}}}
\def\tN{{\tens{N}}}
\def\tO{{\tens{O}}}
\def\tP{{\tens{P}}}
\def\tQ{{\tens{Q}}}
\def\tR{{\tens{R}}}
\def\tS{{\tens{S}}}
\def\tT{{\tens{T}}}
\def\tU{{\tens{U}}}
\def\tV{{\tens{V}}}
\def\tW{{\tens{W}}}
\def\tX{{\tens{X}}}
\def\tY{{\tens{Y}}}
\def\tZ{{\tens{Z}}}

\def\gA{{\mathcal{A}}}
\def\gB{{\mathcal{B}}}
\def\gC{{\mathcal{C}}}
\def\gD{{\mathcal{D}}}
\def\gE{{\mathcal{E}}}
\def\gF{{\mathcal{F}}}
\def\gG{{\mathcal{G}}}
\def\gH{{\mathcal{H}}}
\def\gI{{\mathcal{I}}}
\def\gJ{{\mathcal{J}}}
\def\gK{{\mathcal{K}}}
\def\gL{{\mathcal{L}}}
\def\gM{{\mathcal{M}}}
\def\gN{{\mathcal{N}}}
\def\gO{{\mathcal{O}}}
\def\gP{{\mathcal{P}}}
\def\gQ{{\mathcal{Q}}}
\def\gR{{\mathcal{R}}}
\def\gS{{\mathcal{S}}}
\def\gT{{\mathcal{T}}}
\def\gU{{\mathcal{U}}}
\def\gV{{\mathcal{V}}}
\def\gW{{\mathcal{W}}}
\def\gX{{\mathcal{X}}}
\def\gY{{\mathcal{Y}}}
\def\gZ{{\mathcal{Z}}}

\def\sA{{\mathbb{A}}}
\def\sB{{\mathbb{B}}}
\def\sC{{\mathbb{C}}}
\def\sD{{\mathbb{D}}}
\def\sF{{\mathbb{F}}}
\def\sG{{\mathbb{G}}}
\def\sH{{\mathbb{H}}}
\def\sI{{\mathbb{I}}}
\def\sJ{{\mathbb{J}}}
\def\sK{{\mathbb{K}}}
\def\sL{{\mathbb{L}}}
\def\sM{{\mathbb{M}}}
\def\sN{{\mathbb{N}}}
\def\sO{{\mathbb{O}}}
\def\sP{{\mathbb{P}}}
\def\sQ{{\mathbb{Q}}}
\def\sR{{\mathbb{R}}}
\def\sS{{\mathbb{S}}}
\def\sT{{\mathbb{T}}}
\def\sU{{\mathbb{U}}}
\def\sV{{\mathbb{V}}}
\def\sW{{\mathbb{W}}}
\def\sX{{\mathbb{X}}}
\def\sY{{\mathbb{Y}}}
\def\sZ{{\mathbb{Z}}}

\def\emLambda{{\Lambda}}
\def\emA{{A}}
\def\emB{{B}}
\def\emC{{C}}
\def\emD{{D}}
\def\emE{{E}}
\def\emF{{F}}
\def\emG{{G}}
\def\emH{{H}}
\def\emI{{I}}
\def\emJ{{J}}
\def\emK{{K}}
\def\emL{{L}}
\def\emM{{M}}
\def\emN{{N}}
\def\emO{{O}}
\def\emP{{P}}
\def\emQ{{Q}}
\def\emR{{R}}
\def\emS{{S}}
\def\emT{{T}}
\def\emU{{U}}
\def\emV{{V}}
\def\emW{{W}}
\def\emX{{X}}
\def\emY{{Y}}
\def\emZ{{Z}}
\def\emSigma{{\Sigma}}

\newcommand{\etens}[1]{\mathsfit{#1}}
\def\etLambda{{\etens{\Lambda}}}
\def\etA{{\etens{A}}}
\def\etB{{\etens{B}}}
\def\etC{{\etens{C}}}
\def\etD{{\etens{D}}}
\def\etE{{\etens{E}}}
\def\etF{{\etens{F}}}
\def\etG{{\etens{G}}}
\def\etH{{\etens{H}}}
\def\etI{{\etens{I}}}
\def\etJ{{\etens{J}}}
\def\etK{{\etens{K}}}
\def\etL{{\etens{L}}}
\def\etM{{\etens{M}}}
\def\etN{{\etens{N}}}
\def\etO{{\etens{O}}}
\def\etP{{\etens{P}}}
\def\etQ{{\etens{Q}}}
\def\etR{{\etens{R}}}
\def\etS{{\etens{S}}}
\def\etT{{\etens{T}}}
\def\etU{{\etens{U}}}
\def\etV{{\etens{V}}}
\def\etW{{\etens{W}}}
\def\etX{{\etens{X}}}
\def\etY{{\etens{Y}}}
\def\etZ{{\etens{Z}}}

\newcommand{\pdata}{p_{\rm{data}}}
\newcommand{\ptrain}{\hat{p}_{\rm{data}}}
\newcommand{\Ptrain}{\hat{P}_{\rm{data}}}
\newcommand{\pmodel}{p_{\rm{model}}}
\newcommand{\Pmodel}{P_{\rm{model}}}
\newcommand{\ptildemodel}{\tilde{p}_{\rm{model}}}
\newcommand{\pencode}{p_{\rm{encoder}}}
\newcommand{\pdecode}{p_{\rm{decoder}}}
\newcommand{\precons}{p_{\rm{reconstruct}}}

\newcommand{\laplace}{\mathrm{Laplace}} 

\newcommand{\E}{\mathbb{E}}
\newcommand{\Ls}{\mathcal{L}}
\newcommand{\R}{\mathbb{R}}
\newcommand{\emp}{\tilde{p}}
\newcommand{\lr}{\alpha}
\newcommand{\reg}{\lambda}
\newcommand{\rect}{\mathrm{rectifier}}
\newcommand{\softmax}{\mathrm{softmax}}
\newcommand{\sigmoid}{\sigma}
\newcommand{\softplus}{\zeta}
\newcommand{\KL}{D_{\mathrm{KL}}}
\newcommand{\Var}{\mathrm{Var}}
\newcommand{\standarderror}{\mathrm{SE}}
\newcommand{\Cov}{\mathrm{Cov}}
\newcommand{\normlzero}{L^0}
\newcommand{\normlone}{L^1}
\newcommand{\normltwo}{L^2}
\newcommand{\normlp}{L^p}
\newcommand{\normmax}{L^\infty}

\newcommand{\parents}{Pa} 

\let\ab\allowbreak

\section{Models}
We will first introduce our baseline model and its integration for multiple classification tasks in a multi-task learning (MTL) setup. Next, we will introduce our \modelname{} framework, an automatic way of selecting auxiliary tasks and learning their optimal training mixing ratio w.r.t. the primary task, via a Beta-Bernoulli bandit with Thompson Sampling and a Gaussian Process framework.

\subsection{Bi-Text Classification Model}
\label{subsec:baseline}
\begin{figure}[t]
\centering
\includegraphics[width=0.9\linewidth]{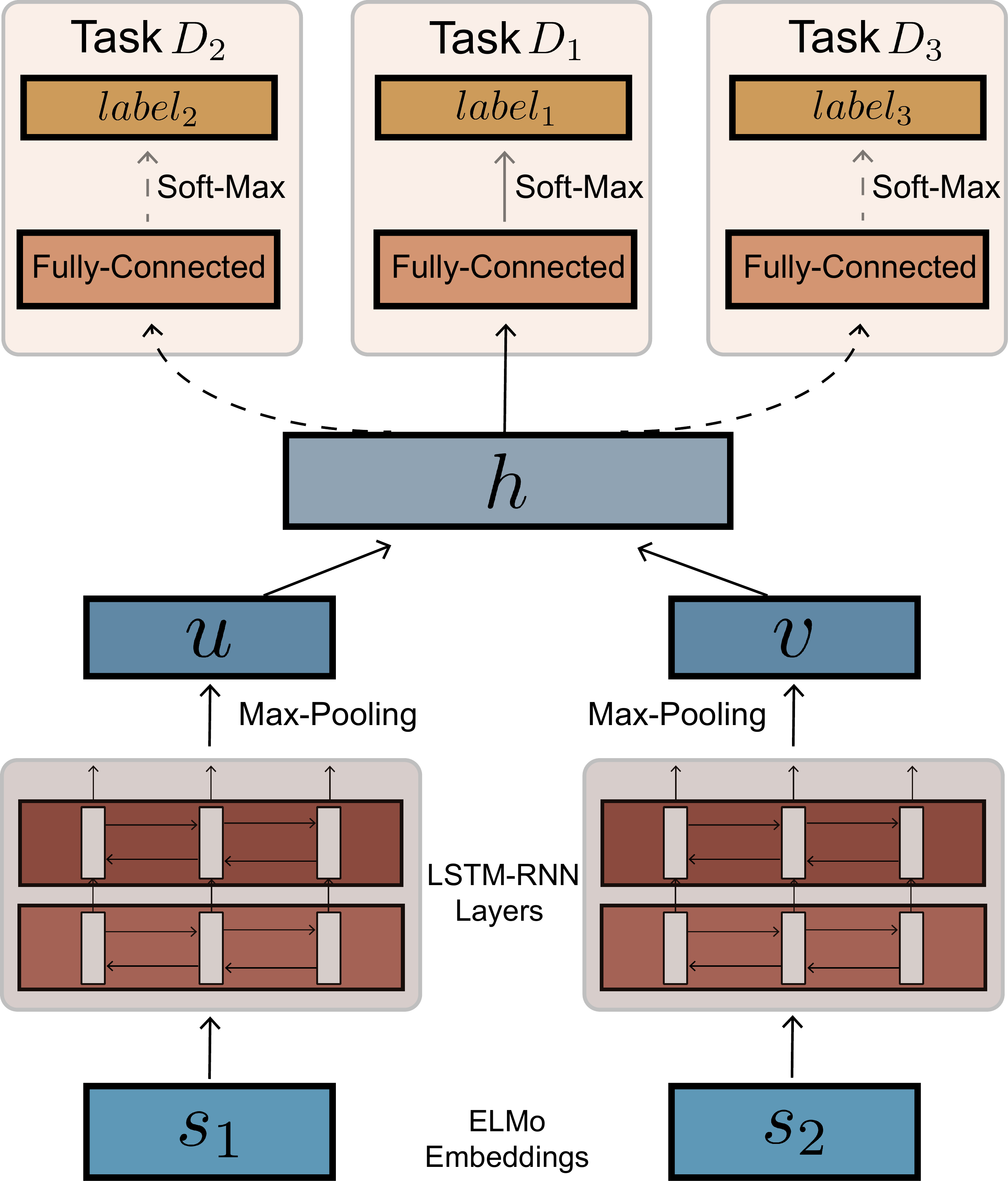}
\caption{Overview of our baseline model where we use different projection layers for each task during MTL, while sharing rest of the model parameters.
\label{fig:baseline}
\vspace{-10pt}
}
\end{figure}

Let $\vs_1$ and $\vs_2$ be the input sentence pair in our classification task, where we encode these sentences via bidirectional LSTM-RNN, similar to that of~\newcite{conneau2017supervised}. Next, we do max-pooling on the output hidden states of both encoders where $\vu$ and $\vv$ are the outputs from the max-pooing layer for $\vs_1$ and $\vs_2$ respectively. Later, we map these two representations ($\vu$ and $\vv$) into a single rich dense representation vector $\vh$:
\begin{equation}
    \vh = [\vu ; \vv; \vu \star \vv; |\vu-\vv|]
\end{equation}
where $[;]$ represents the concatenation and $\vu\star \vv$ represents the element-wise multiplication of $\vu$ and $\vv$. We project this final representation $\vh$ to label space to classify the given sentence pair (see Fig.~\ref{fig:baseline}). We also use ELMo~\cite{peters2018Deep} representations for word embeddings in our model. For this, we extract the three ELMo layer representations for each of the sentence pair and use their weighted sum as the ELMo output representation, where the weights are trainable.

\begin{figure*}[t]
\centering
\includegraphics[width=0.9\linewidth]{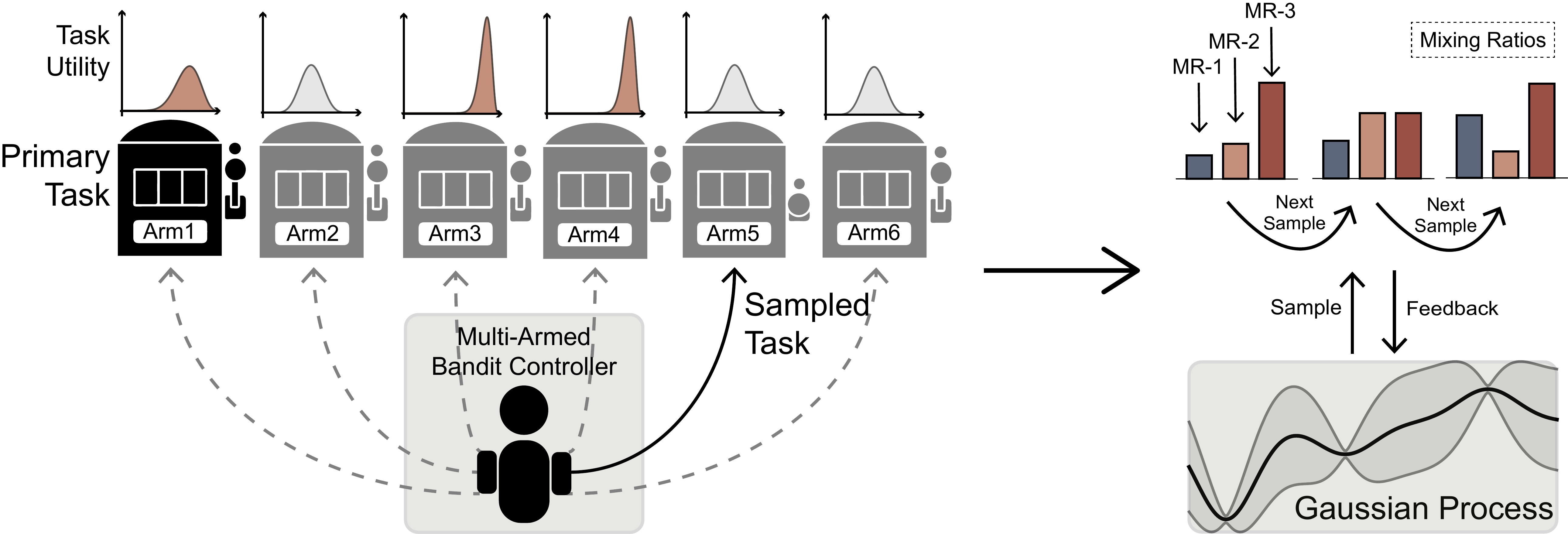}
\vspace{-5pt}
\caption{Overview of our \modelname{} framework. \textbf{Left}: the multi-armed bandit controller used for task selection, where each arm represents a candidate auxiliary task. The agent iteratively pulls an arm, observes a reward, updates its estimates of the arm parameters, and samples the next arm. \textbf{Right}: the Gaussian Process controller used for automatic mixing ratio (MR) learning. The GP controller sequentially makes a choice of mixing ratio, observes a reward, updates its estimates, and selects the next mixing ratio to try, based on the full history of past observations.
\label{fig:mtl}
}
\vspace{-15pt}
\end{figure*}

\subsection{Multi-Task Learning}
\label{subsec:mtl-intro}

In this work, we focus on improving a task (primary task) by allowing it to share parameters with related auxiliary tasks via multi-task learning (MTL). Let $\{D_1, ..., D_N\}$ be a set of $N$ tasks, where we set $D_1$ to be the primary task and the rest of them as auxiliary tasks. We can extend our single-task learning baseline (see Sec.~\ref{subsec:baseline}) into multi-task learning model by augmenting the model with $N$ projection layers while sharing the rest of the model parameters across these $N$ tasks (see Fig.~\ref{fig:baseline}). We employ MTL training of these tasks in alternate mini-batches based on a mixing ratio $\eta_1{:}\eta_2{:}..\eta_N$, similar to previous work~\cite{luong2015multi}, where we optimize $\eta_i$ mini-batches of task $i$ and go to the next task.

In MTL, choosing the appropriate auxiliary tasks and properly tuning the mixing ratio can be important for the performance of multi-task models.
The naive way of trying all combinations of task selections is hardly tractable.
To solve this issue, we propose \modelname{}, a two-stage pipeline in the next section. In the first stage, we automatically find the relevant auxiliary tasks (out of the given $N-1$ options) which improve the performance of the primary task. After finding the relevant auxiliary tasks, in the second stage, we take these selected tasks along with the primary task and automatically learn their training mixing ratio.

\subsection{Automatic Task Selection: Multi-Armed Bandit with Thompson Sampling}
\label{subsec:task-selection}
Tuning the mixing ratio for $N$ tasks in MTL becomes exponentially harder as the number of auxiliary tasks grows very large. However, in most circumstances, only a small number of these auxiliary tasks are useful for improving the primary task at hand. Manually searching for this optimal choice of relevant tasks is intractable. Hence, in this work, we present a method for automatic task selection via multi-armed bandits with Thompson Sampling (see the left side of Fig.~\ref{fig:mtl}).

Let $\{a_1,...,a_N\}$ represent the set of $N$ arms (corresponding to the set of tasks $\{D_1,..., D_N\}$) of the bandit controller in our multi-task setting, where the controller selects a sequence of actions/arms over the current training trajectory to maximize the expected future payoff. At each round $t_b$, the controller selects an arm based on the noisy value estimates and observes rewards $r_{t_b}$ for the selected arm.
Let $\theta_k \in [0, 1]$ be the utility (usefulness) of task $k$. Initially, the agent begins with an independent prior belief over $\theta_k$. We take these priors to be Beta-distributed with parameters $\alpha_k$ and $\beta_k$, and the prior probability density function of $\theta_k$ is:
\begin{equation}
p(\theta_k) = \frac{\Gamma(\alpha_k+\beta_k)}{\Gamma(\alpha_k)\Gamma(\beta_k)} \theta_k^{\alpha_k-1} (1-\theta_k)^{\beta_k-1}
\end{equation}
\label{eq:beta}
where $\Gamma$ denotes the gamma function. We formulate the reward $r_{t_b} \in \{0,1\}$ at round $t_b$ as a Bernoulli variable, where an action $k$ produces a reward of $1$ with a chance of $\theta_k$ and a reward of $0$ with a chance of $1-\theta_k$. The true utility of task $k$, i.e., $\theta_k$, is unknown, and may or may not change over time (based on stationary vs. non-stationary of task utility). We define the reward as whether sampling the task $k$ improves (or maintains) the validation metric of the primary task,

\begin{equation}
r_{t_b}=
\begin{cases}
	1, & \text{if}\ R_{t_b} \geq R_{t_b - 1} \\
	0, & \text{otherwise}
\end{cases}
\end{equation}
\label{eq:initial-binary-reward}

where $R_{t_b}$ represents the validation performance of the primary task at time $t_b$. With our reward setup above, the utility of each task ($\theta_k$) can be intuitively interpreted as the probability that multi-task learning with task $k$ can improve (or maintain) the performance of the primary task. The conjugacy properties of the Beta distribution assert that the posterior distribution is also Beta with parameters that can be updated using a simple Bayes rule, which is defined as follows~\cite{russo2018tutorial},
\begin{equation}
\begin{split}
p(\theta_k \lvert r)
	& \propto \text{Bern}_{\theta}(r) \text{Beta}_{\alpha, \beta}(\theta_k) \\
    & \propto \text{Beta}_{\alpha + r, \beta + 1 - r}(\theta_k)
\end{split}
\end{equation}
\vspace{-10pt}
\begin{equation}
(\alpha_k, \beta_k) =
\begin{cases}
	(\alpha_k, \beta_k), & \hspace{-6pt} \text{if}\ x^s_{t_b} \neq k  \\
	(\alpha_k, \beta_k) \mbox{+} (r_{t_b}, 1 - r_{t_b}), & \hspace{-6pt} \text{if}\ x^s_{t_b} = k
\end{cases}
\end{equation}
\label{eq:stationary-posterior-update}
where $x^s_{t_b}$ is the sampled task at round $t_{b}$. Finally, at the end of the training, we calculate the expected value of each arm as follows:
\begin{equation}
\mathbb{E}_p [\theta_k] = \frac{\alpha_k}{\alpha_k + \beta_k}
\end{equation}
\label{eq:utility}
Here, the expectation measures the probability of improving (or maintaining) the primary task by sampling this task. To decide the next action to take, we apply Thompson Sampling~\citep{russo2018tutorial,chapelle2011empirical} to trade off exploitation (maximizing immediate performance) and exploration (investing to accumulate new information that might improve performance in the future). In Thompson Sampling~\cite{russo2018tutorial}, instead of taking action $k$ that maximizes the expectation (i.e., $\arg\max_k \mathbb{E}_p [\theta_k]$), we randomly sample the primary task improvement probability $\hat{\theta}_k$ from the posterior distribution $\hat{\theta}_k \sim p(\theta_k)$, and take the action $k$ that maximizes the sampled primary task improvement probability, i.e., $\arg\max_k \hat{\theta}_k$.
At the end of the training, the task selection can proceed either via a threshold on the expectation, or take the top-$K$ tasks, and run stage-2 using the selected task subset as auxiliary tasks (details in Sec.~\ref{subsec:gp}).

\paragraph{Stronger Prior for Primary Task} Note that at the beginning of training, model performance is usually guaranteed to improve from the initial random choices. This causes issues in updating arm values because less useful tasks will be given high arm values when they happen to be sampled at the beginning. To resolve this issue, we initially set a slightly stronger prior/arm-value in favor of the arm corresponding to the primary task. Intuitively, the bandit will then sample the primary model more often at the beginning, and then start exploring auxiliary tasks when the primary model's performance stabilizes (as the arm value of the primary model will start decreasing because sampling it in later rounds produces smaller additional improvements). 

\paragraph{Non-Stationary Multi-Armed Bandit} Also note that the intrinsic usefulness of each task varies throughout the training (e.g., the primary task might be more important at the beginning, but not necessarily at the end), and thus the agent faces a non-stationary system. In such cases, the agent should always be encouraged to explore in order to track changes as the system drifts. One simple approach to inject non-stationarity is to discount the relevance of previous observations. Thus we introduce a tunable decay ratio $\gamma$, and modify Eq.~\ref{eq:stationary-posterior-update} as follows:
\begin{equation}
(\alpha_k, \beta_k) =
\begin{cases}
	(\hat{\alpha}_k, \hat{\beta}_k), & \hspace{-6pt} \text{if}\ k \neq x^s_{t_b}  \\
	(\hat{\alpha}_k, \hat{\beta}_k) \mbox{+} (r_{t_b}, 1 - r_{t_b}), & \hspace{-6pt} \text{if}\ k = x^s_{t_b}
\end{cases}
\end{equation}
where $\hat{\alpha}_k = (1-\gamma) \alpha_k + \gamma \alpha_0$ and $\hat{\beta}_k = (1-\gamma) \beta_k + \gamma \beta_0$, and $\gamma$ controls how quickly uncertainty is injected into the system ($\alpha_0, \beta_0$ are parameters of the prior). Algorithm~\ref{alg:BernoulliTS} presents the Thompson Sampling algorithm with a Beta-Bernoulli MAB.

\begin{algorithm}[t]
\begin{small}
\caption{$\text{BernThompson}(N, \alpha, \beta, \gamma, \alpha_0, \beta_0)$}\label{alg:BernoulliTS}
\begin{algorithmic}[1]
\For{$t_b=1,2,\ldots $}
\State \# sample model:
\For{$k=1, \ldots, N$}
\State Sample $\hat{\theta}_k \sim \text{Beta}(\alpha_k, \beta_k)$
\EndFor
\State \# select and apply action:
\State $x^s_{t_b} \leftarrow \arg\max_k \hat{\theta}_k$
\State Apply $x^s_{t_b}$ and observe $r_{t_b}$ 
\State \# non-stationarity
\For{$k=1, \ldots, N$}
\State $\hat{\alpha}_k = (1-\gamma) \alpha_k + \gamma \alpha_0$
\State $\hat{\beta}_k = (1-\gamma) \beta_k + \gamma \beta_0$ 
\If {$k \neq x^s_{t_b}$}
\State $(\alpha_{k}, \beta_{k}) \leftarrow  (\hat{\alpha}_k, \hat{\beta}_k)$
\Else
\State $(\alpha_{k}, \beta_{k}) \leftarrow  (\hat{\alpha}_k, \hat{\beta}_k) \mbox{+} (r_{t_b}, 1-r_{t_b})$
\EndIf
\EndFor 
\EndFor
\end{algorithmic}
\end{small}
\end{algorithm}

\subsection{Automatic Mixing Ratio Learning via Gaussian Process}
\label{subsec:gp}
The right side of Fig.~\ref{fig:mtl} illustrates our Gaussian Process controller for automatic learning of the MTL training mixing ratio (see definition in Sec.~\ref{subsec:mtl-intro}). Given the selected auxiliary tasks from the previous section, the next step is to find a proper mixing ratio of training these selected tasks along with the primary task.\footnote{Note that ideally Gaussian Process can also learn to set the mixing ratio of less important tasks to zero, hence allowing it to essentially also perform the task selection step. However, in practice, first applying our task selection Thompson-Sampling model (Sec.~\ref{subsec:task-selection}) allows GP to more efficiently search the mixing ratio space for the small number of filtered auxiliary tasks, as shown in results of Sec.~\ref{subsec:analysis}.}
Manual tuning of this mixing ratio via a large grid search over the hyperparameter values is very time and compute expensive (even when the number of selected auxiliary tasks is small, e.g., 2 or 3). Thus, in our second stage, we instead apply a non-parametric Bayesian approach to search for the approximately-optimal mixing ratio. In particular, we use a `Gaussian Process' to sequentially search for the mixing ratio by trading off exploitation and exploration automatically.
Next, we describe our Gaussian Process approach in detail.

A Gaussian Process~\cite{rasmussen2004gaussian,snoek2012practical,shahriari2016taking}, $\text{GP}(\mu_0, k)$, is a non-parametric model that is fully characterized by a mean function $\mu_0: \mathcal{X} \mapsto \mathbb{R}$ and a positive-definite kernel or covariance function $k: \mathcal{X} \times \mathcal{X} \mapsto \mathbb{R}$. Let $\vx_1, \vx_2, ..., \vx_n$ denote any finite collections of $n$ points, where each $\vx_i$ represents a choice of the mixing ratio (i.e., the ratio $\eta_1{:}\eta_2{:}..\eta_N$ described in Sec.~\ref{subsec:mtl-intro}), and $f_i = f(\vx_i)$ is the (unknown) function values evaluated at $\vx_i$ (true performance of the model given the selected mixing ratio). Let $y_1, y_2, ..., y_n$ be the corresponding noisy observations (the validation performance at the end of training). In the context of GP Regression (GPR), $\vf =\{f_1, ..., f_n\}$ are assumed to be jointly Gaussian~\cite{rasmussen2004gaussian}, i.e., $\vf \lvert \mX \sim \mathcal{N}(\vm, \mK)$, where, $\vm_i = \mu_0(\vx_i)$ is the mean vector, and $\mK_{i,j} = k(\vx_i, \vx_j)$ is the covariance matrix. Then the noisy observations $\vy = y_1, ..., y_n$ are normally distributed around $\vf$ as follows: $\vy \lvert \vf \sim \mathcal{N}(\vf, \sigma^2 \mI)$.

Given $\mathcal{D}=(\vx_1, y_1), ..., (\vx_{n_0}, y_{n_0})$, the set of random initial observations, where $\vx_i$ represents a mixing ratio and $y_i$ represents the corresponding model's validation performance. Next, we model the GP based on these initial observations as described above. We sample a next point $\vx_{n_0+1}$ (a mixing ratio in our case) from this GP and get its corresponding model performance $y_{n_0+1}$, and update the GP again by now considering the $n_0+1$ points~\cite{rasmussen2004gaussian}. We continue this process for a fixed number of steps.  Next, we will discuss how we perform the sampling (based on acquisition functions) and the kernels used for calculating the covariance.

\begin{table*}
\begin{center}
\begin{small}
\begin{tabular}{|l|c|c|c|c|c|}
\hline
Models & RTE & MRPC & QNLI & CoLA & SST-2 \\
\hline
BiLSTM+ELMo (Single-Task) ~\cite{wang2018glue} & 50.1 & 69.0/80.8 & 69.4 & \textbf{35.0} & 90.2 \\
BiLSTM+ELMo (Multi-Task) ~\cite{wang2018glue} & 55.7 & 76.2/83.5 & 66.7 & 27.5 & 89.6 \\
\hline
\hline
Our Baseline &
    54.0 & 75.7/83.7 & 74.0 & 30.8 & 91.3 \\
    
Our \modelname{} &
    \textbf{58.7} & \textbf{78.5/84.5} & \textbf{79.2} & 32.9 & \textbf{91.8} \\
    
\hline
\end{tabular}
\end{small}
\end{center}
\vspace{-10pt}
\caption{Test GLUE results of previous work, our baseline, and our \modelname{} MTL framework. We report accuracy and F1 for MRPC, Matthews correlation for CoLA, and accuracy for all others.
\label{table:test-results}
\vspace{-12pt}
}
\end{table*}

\paragraph{Acquisition Functions}
Here, we describe the acquisition functions for deciding where to sample next. While one could select the points that maximize the mean function, this does not always lead to the best outcome~\cite{hoffman2011portfolio}. Since we also have the variance of the estimates along with the mean value of each point $\vx_i$, we can incorporate this information into the optimization.
In this work, we use the GP-Hedge approach~\cite{hoffman2011portfolio,auer1995gambling}, which probabilistically chooses one of three acquisition functions: probability of improvement, expected improvement, and upper confidence bound. Probability of improvement acquisition functions measure the probability that the sampled mixing ratio $\vx_i$ leads to an improvement upon the best observed value so far ($\tau$), $\mathbb{P}(f(\vx_i) > \tau)$. Expected improvement additionally incorporates the amount of improvement, $\mathbb{E}[(f(\vx_i) - \tau)\mathbb{I}(f(\vx_i) > \tau)]$. The Gaussian Process upper confidence bound (GP-UCB) algorithm measures the optimistic performance upper bound of the sampled mixing ratio~\cite{srinivas2009gaussian}, $\mu_i(\vx_i) + \lambda\sigma_i(\vx_i)$, for some hyper-parameter $\lambda$.

\paragraph{Matern Kernel} The covariance function (or kernel) defines the nearness or similarity of two points in the Gaussian Process. Here, we use the automatic relevance determination (ARD) Matern kernel~\cite{rasmussen2004gaussian}, which is parameterized by $\nu > 0$ that controls the level of smoothness. In particular, samples from a GP with such a kernel are differentiable $\floor{\nu - 1}$ times. When $\nu$ is half-integer (i.e. $\nu = p + 1/2$ for non-negative integer $p$), the covariance function is a product of an exponential and a polynomial of order $p$. In the context of machine learning, usual choices of $\nu$ include $3/2$ and $5/2$~\cite{shahriari2016taking}.

\section{Experiment Setup}
\label{sec:setup}

\noindent\textbf{Datasets}:
We evaluate our models on several datasets from the GLUE benchmark~\cite{wang2018glue}: RTE, QNLI, MRPC, SST-2, and CoLA. For all these datasets, we use the standard splits provided by~\newcite{wang2018glue}. For dataset details, we refer the reader to the GLUE paper.\footnote{We did not include the remaining tasks as primary tasks, because STS-B is a regression task; MNLI is a very large dataset and does not benefit much from MTL with other tasks in the GLUE benchmark; and QQP and WNLI have dev/test discrepancies and adversarial label issues as per the GLUE website's FAQ: \url{https://gluebenchmark.com/faq}}

\noindent\textbf{Training Details}:
We use pre-trained ELMo\footnote{\url{https://allennlp.org/elmo}} to obtain sentence representations as inputs to our model~\cite{peters2018Deep}, and the Gaussian Process implementation is based on Scikit-Optimize\footnote{\url{https://scikit-optimize.github.io}}, and we adopt most of the default configurations. We use accuracy as the validation criterion for all tasks. For all of our experiments except QNLI and SST-2, we apply early stopping on the validation performance plateau.\footnote{In our initial experiments, we found early stopping on larger datasets led to sub-optimal performance, and hence we used a pre-specified maximum number of steps instead.} The set of candidate auxiliary tasks consists of all 2-sentence classification tasks when the primary task is a classification of two sentences, whereas it consists of all two-sentence and single-sentence classification tasks when the primary task is a classification of a single sentence.\footnote{We made this design decision because there are only two single-sentence tasks in GLUE, so we mix them with 2-sentence tasks to allow more auxiliary choices.} Since the utility estimates from the multi-armed bandit controller are noisy, we choose the top two tasks based on expected task utility estimates, and include additional tasks if their utility estimate is above 0.5. All the results reported are the aggregate of the same experiment with two runs (with different random seeds) unless explicitly mentioned.\footnote{We use the average of validation results across runs as the tuning criterion, and use the ensemble of models across runs for reporting the test results.} We use a two-layer LSTM-RNN with hidden size of 1024 for RTE and 512 for the rest of the models, and use Adam Optimizer~\cite{Kingma2014AdamAM}. The prior parameters of each task in stage-1 are set to be $\alpha_0=1$, $\beta_0=1$, which are commonly used in other literature. For stage-1, the bandit controller iteratively selects batches of data from different tasks during training to learn the approximate importance of each auxiliary task~\cite{graves2017automated}. In stage-2 (Gaussian Process), we sequentially draw samples of mixing ratios and evaluate each sample after full training~\cite{snoek2012practical}. Without much tuning, we used approximately 200 rounds for the stage-1 bandit-based approach, where each round consist of approximately 10 mini-batches of optimization. For stage-2, we experimented with 15 and 20 as the number of samples to draw and found that 15 samples for MRPC and 20 samples for the rest of the tasks work well. This brings the total computational cost for our two-stage pipeline to be approximately (15+1)x and (20+1)x, where x represents the time taken to run the baseline model for the given task. This is significantly more efficient than a grid-search based manually-tuned mixing ratio setup (which would scale exponentially with the number of tasks).

\section{Results}

\subsection{Baseline Models}
Table~\ref{table:test-results} shows the results of our baseline and previous works~\cite{wang2018glue}. We can see that our single-task baseline models achieve stronger performance on almost all tasks in comparison to previous work's single-task models.\footnote{\label{note1}Note that we do not report previous works which \emph{fine-tune} large external language models for the task (e.g., OpenAI-GPT and BERT), because they are not fairly comparable w.r.t. our models. Similarly, we report the non-attention based best GLUE models (i.e., BiLSTM+ELMo) for a fair comparison to our non-attention baseline. Our approach should ideally scale to large pre-training/fine-tuning models like BERT, given appropriate compute resources.} Next, we present the performance of our \modelname{} framework on top of these strong baselines.

\subsection{Multi-Task Models}
Table~\ref{table:test-results} also presents the performance of our \modelname{} framework-based MTL models. As can be seen, our MTL models improve significantly (see Table~\ref{table:validation-results} for standard deviations) upon their corresponding single-task baselines for all tasks, and achieve strong improvements as compared to the fairly-comparable\textsuperscript{\ref{note1}} multi-task results of previous work~\cite{wang2018glue}.\footnote{Note that even though the performance improvement gaps of~\newcite{wang2018glue} (MTL vs. baseline) and our improvements (\modelname{} vs. our improved baseline) are similar, these are inherently two different setups. \newcite{wang2018glue} MTL is based on a `one model for all' setup~\cite{kaiser2017one,mccann2018natural}, whereas our approach interpretably chooses the 2-3 tasks that are most beneficial for the given primary task. Also see Sec.~\ref{sec:setup} for comparison of training speeds for these two setups.} 
During the task selection stage of our \modelname{} framework, we observe that MultiNLI is chosen as one of the auxiliary tasks in all of our MTL models. This is intuitive given that MultiNLI contains multiple genres covering diverse aspects of the complexity of language~\cite{conneau2017supervised}. Also, we observe that WNLI is sometimes chosen in the task selection stage; however, it is always dropped (mixing ratio of zero) by the Gaussian Process controller, showing that it is not beneficial to use WNLI as an auxiliary task (intuitive, given its small size). Next, we discuss the improvements on each of the primary tasks and the corresponding auxiliary tasks selected by \modelname{} framework.

\noindent\textbf{RTE}: 
Our \modelname{} approach achieves stronger results w.r.t. the baseline on RTE (58.7 vs. 54.0). During our task selection stage, we found out that QQP and MultiNLI tasks are important for RTE as auxiliary tasks. For the second stage of automatic mixing ratio learning via Gaussian Process, the model learns that a mixing ratio of 1:5:5 works best to improve the primary task (RTE) using related auxiliary tasks of QQP and MultiNLI.

\noindent\textbf{MRPC}:
\modelname{} here performs much better than the baseline on MRPC (78.5/84.5 vs. 75.7/83.7). During our task selection stage, we found out that RTE and MultiNLI tasks are important for MRPC as auxiliary tasks. In the second stage,  \modelname{} learned a mixing ratio of 9:1:4 for these three tasks (MRPC:RTE:MultiNLI).

\noindent\textbf{QNLI}:
Again, we achieve substantial improvements with \modelname{} w.r.t. baseline on QNLI (79.2 vs. 74.0). Our task selection stage learned that WNLI and MultiNLI tasks are best as auxiliary tasks for QNLI. 
We found that the Gaussian Process further drops WNLI by setting its mixing ratio to zero, and returns 20:0:5 as the best mixing ratio for QNLI:WNLI:MultiNLI.

\noindent\textbf{CoLA}:
We also observe a strong performance improvement on CoLA with our \modelname{} model w.r.t. our baseline (32.9 vs. 30.8). During our task selection stage, we found out that MultiNLI and WNLI tasks are important for CoLA as auxiliary tasks. In the second stage, GP learns to drop WNLI, and found the mixing ratio of 20:5:0 for CoLA:MultiNLI:WNLI.

\noindent\textbf{SST-2}:
Here also our \modelname{} approach performs better than the baseline (91.8 vs. 91.3). The task selection stage chooses MultiNLI, MRPC, and WNLI as auxiliary tasks and the stage-2 Gaussian Process model drops MRPC and WNLI by setting their mixing ratio to zero (learns ratio of 13:5:0:0 for SST-2:MultiNLI:MRPC:WNLI).

\begin{table}
\centering
\begin{tabular}{|l|c|c|}
	\hline
	Name & Validation & Test \\
	\hline
    Baseline     & 78.3 & 75.7/83.7 \\
    w/o Stage-1  & 80.3 & 76.3/83.8 \\
    w/o Stage-2  & 80.3 & 76.7/83.8 \\
    Final MTL    & 81.2 & 78.5/84.5 \\
	\hline
\end{tabular}
\vspace{-7pt}
\caption{Ablation results on the two stages of our \modelname{} framework on MRPC.
\label{table:ablation-results}
\vspace{-9pt}}
\end{table}

\section{Analysis}
\subsection{Ablation on MTL stages}
\label{subsec:analysis}

In this section, we examine the usefulness of each stage of our two-stage MTL pipeline.\footnote{We present this ablation only on MRPC for now, because GP stage-2 takes a lot of time without the task selection stage.}

\noindent\textbf{Removing Stage-1}:
The purpose of the Beta-Bernoulli MAB in stage-1 is to find useful auxiliary tasks for the given primary task. Here, to understand its importance, we remove the task selection part, and instead directly run the Gaussian Process (GP) model on all tasks (see `w/o Stage-1' row in Table~\ref{table:ablation-results}). We can see that by removing the task selection stage, the Gaussian Process model can still outperform the baseline, indicating the usefulness of the GP, but the large mixing ratio search space causes the GP to be unable to efficiently find the best mixing ratio setting.

\noindent\textbf{Removing Stage-2}:
Given the selected tasks from stage-1, the goal of the Gaussian Process in stage-2 is to efficiently find the approximately-optimal mixing ratio. To examine its usefulness, we replace the Gaussian Process controller by manually tuning a grid of mixing ratios, where the number of tuning experiments equals to the number of steps used in the Gaussian Process model (for a fair comparison). Table~\ref{table:ablation-results} shows the results by removing stage-2. We can see that a grid search over hyper-parameters can improve upon the baseline, indicating the usefulness of stage-1 task selection, but a reasonable-sized fair-comparison grid search (i.e., not exhaustive over all ratio values) is not able to match our stage-2 GP process that leverages prior experimental results to more efficiently find the best setting.

\subsection{Stability of MTL Models}
In this section, we provide the mean and standard deviation of our baseline and multi-task models (over three runs) on the validation set. Note that the test set is hidden, so we cannot do these studies on it. As seen in Table~\ref{table:validation-results}, our multi-task models clearly surpass the performance of baseline models w.r.t. standard deviation gaps, in all tasks.

\begin{figure}[t]
\centering
\includegraphics[width=0.95\linewidth]{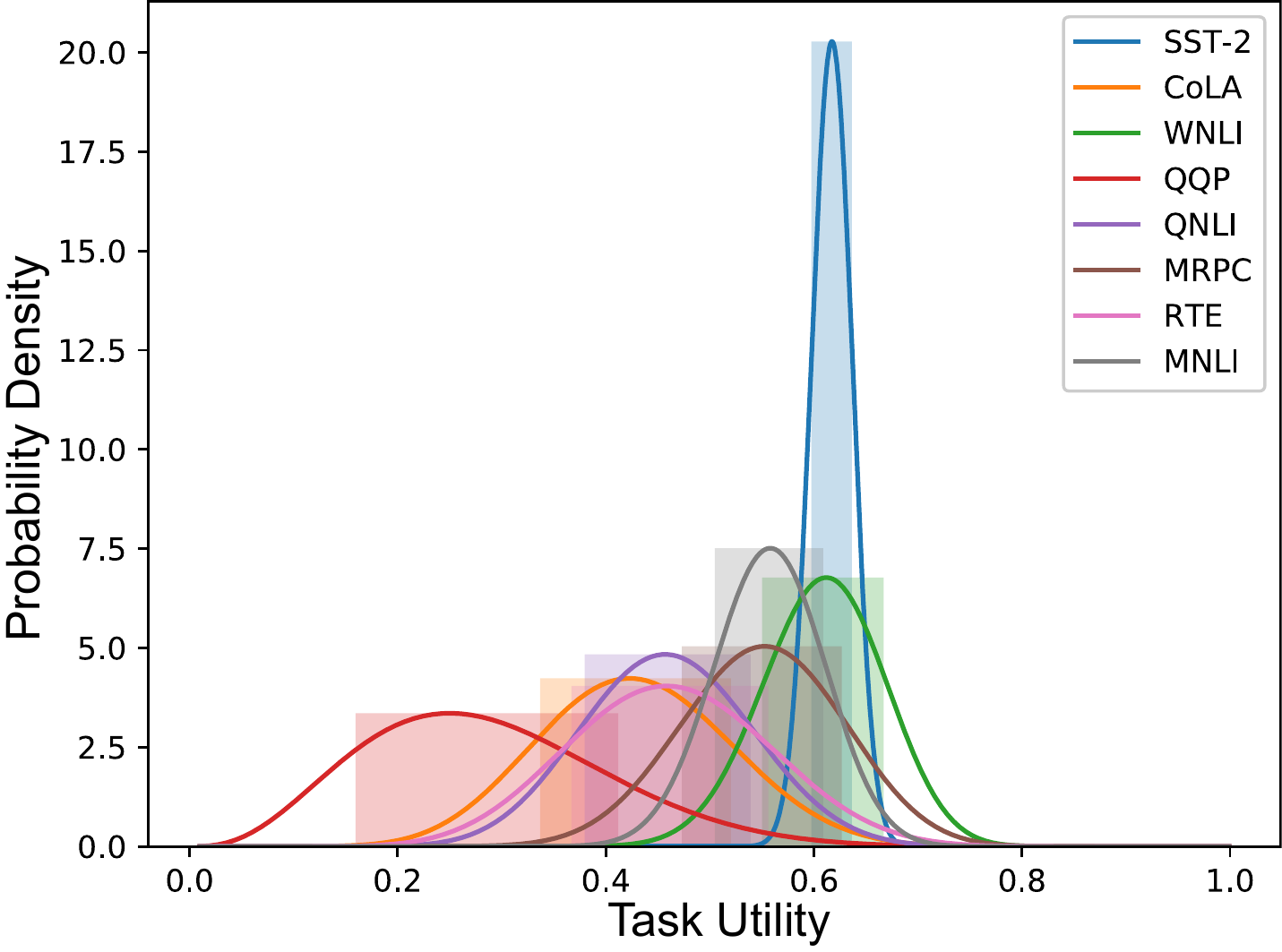}
\vspace{-4pt}
\caption{Visualization of task utility estimates from the multi-armed bandit controller on SST-2 (primary task). The x-axis represents the task utility, and the y-axis represents the corresponding probability density. Each curve corresponds to a task and the bar corresponds to their confidence interval.
\label{fig:visualization}
\vspace{-1pt}
}
\end{figure}

\subsection{Visualization of Task Selection}
In Fig.~\ref{fig:visualization}, we show an example of the task utility estimates from the stage-1 multi-armed bandit controller (Eq.~\ref{eq:beta}) on SST-2. The x-axis represents the task utility, and the y-axis represents the probability density over task utility. Each curve represents a task (the blue curve corresponds to the primary task, SST-2, and the rest of the curves correspond to auxiliary tasks), and the width of the bars represents the confidence interval of their estimates. We can see that the bandit controller gives the highest (and most confident) utility estimate for the primary task, which is intuitive given that the primary task should be the most useful task for learning itself. Further, it gives 2-3 tasks moderate utility estimates (the corresponding expected values are around 0.5), and relatively lower utility estimates for the remaining tasks (the corresponding expected values are lower than 0.5).

\subsection{Educated-Guess Baselines}
We additionally experimented with `educated-guess' baseline models, where MTL is performed using manual intuition mixtures that seem a priori sensible.\footnote{These educated-guess models replace our stage-1 automatic auxiliary task section with manual intuition task-mixtures; but we still use our stage-2 Gaussian Process for mixing ratio learning, for fair comparison.}  For example, with MRPC as the primary task, our first educated-guess baseline is to choose other similar paraphrasing-based auxiliary tasks, i.e., QQP in case of GLUE. This MRPC+QQP model achieves 80.8, whereas our \modelname{} framework chose MRPC+RTE+MultiNLI and achieved 81.2. Furthermore, as our second educated-guess baseline, we added MultiNLI as an auxiliary task (in addition to QQP), since MultiNLI was helpful for all tasks in our MTL experiments. This educated-guess MRPC+QQP+MultiNLI model achieves 80.9 (vs. 81.2 for our \modelname{} model). This suggests that our \modelname{} framework (that automatically chose the seemingly less-related RTE task for MRPC) is equal or better than manual intuition based educated-guess models.

\begin{table}
\small
\centering
\begin{tabular}{|l|c|c|c|c|c|c|}
	\hline
	Name     & RTE  & MRPC & QNLI & CoLA & SST-2 \\
	\hline
	\multicolumn{6}{|c|}{\textsc{Baselines}} \\
	\hline
    Mean & 58.6 & 78.3 & 74.9 & 74.6 & 91.4 \\
    Std  & 0.94 & 0.31 & 0.30 & 0.44 & 0.36 \\
    \hline
	\multicolumn{6}{|c|}{\textsc{Multi-Task Models}} \\
	\hline
    Mean & 62.0 & 81.1 & 76.0 & 75.7 & 91.8 \\
    Std  & 0.62 & 0.20 & 0.18 & 0.18 & 0.29 \\
	\hline
\end{tabular}
\vspace{-5pt}
\caption{Validation-set performance mean and standard deviation (based on three runs) of our baselines and Multi-task models in accuracy.
\label{table:validation-results}
\vspace{-13pt}}

\end{table}

\section{Conclusion}
We presented the \modelname{} framework, a two-stage multi-task learning pipeline, where the first stage automatically selects the relevant auxiliary tasks for the given primary task and the second stage automatically learns their optimal mixing ratio. We showed that \modelname{} performs better than strong baselines on several GLUE tasks. Further, we ablated the importance of each stage of our \modelname{} framework and also discussed the intuition of selected auxiliary tasks.

\section*{Acknowledgments}
We thank the reviewers for their helpful comments. This work was supported by DARPA (YFA17-D17AP00022), ONR (N00014-18-1-2871), Google, Facebook, Baidu, Salesforce, and Nvidia.  
The views contained in this article are those of the authors and not of the funding agency.

\bibliography{main}

\begin{thebibliography}{54}
\expandafter\ifx\csname natexlab\endcsname\relax\def\natexlab#1{#1}\fi

\bibitem[{Auer et~al.(2002{\natexlab{a}})Auer, Cesa-Bianchi, and
  Fischer}]{auer2002finite}
Peter Auer, Nicolo Cesa-Bianchi, and Paul Fischer. 2002{\natexlab{a}}.
\newblock Finite-time analysis of the multiarmed bandit problem.
\newblock \emph{Machine learning}, 47(2-3):235--256.

\bibitem[{Auer et~al.(1995)Auer, Cesa-Bianchi, Freund, and
  Schapire}]{auer1995gambling}
Peter Auer, Nicolo Cesa-Bianchi, Yoav Freund, and Robert~E Schapire. 1995.
\newblock Gambling in a rigged casino: The adversarial multi-armed bandit
  problem.
\newblock In \emph{focs}, page 322. IEEE.

\bibitem[{Auer et~al.(2002{\natexlab{b}})Auer, Cesa-Bianchi, Freund, and
  Schapire}]{auer2002nonstochastic}
Peter Auer, Nicolo Cesa-Bianchi, Yoav Freund, and Robert~E Schapire.
  2002{\natexlab{b}}.
\newblock The nonstochastic multiarmed bandit problem.
\newblock \emph{SIAM journal on computing}, 32(1):48--77.

\bibitem[{Ben-David and Schuller(2003)}]{ben2003exploiting}
Shai Ben-David and Reba Schuller. 2003.
\newblock Exploiting task relatedness for multiple task learning.
\newblock In \emph{Learning Theory and Kernel Machines}, pages 567--580.
  Springer.

\bibitem[{Bingel and S{\o}gaard(2017)}]{bingel2017identifying}
Joachim Bingel and Anders S{\o}gaard. 2017.
\newblock Identifying beneficial task relations for multi-task learning in deep
  neural networks.
\newblock \emph{arXiv preprint arXiv:1702.08303}.

\bibitem[{Brochu et~al.(2010)Brochu, Cora, and De~Freitas}]{brochu2010tutorial}
Eric Brochu, Vlad~M Cora, and Nando De~Freitas. 2010.
\newblock A tutorial on bayesian optimization of expensive cost functions, with
  application to active user modeling and hierarchical reinforcement learning.
\newblock \emph{arXiv preprint arXiv:1012.2599}.

\bibitem[{Bubeck et~al.(2012)Bubeck, Cesa-Bianchi et~al.}]{bubeck2012regret}
S{\'e}bastien Bubeck, Nicolo Cesa-Bianchi, et~al. 2012.
\newblock Regret analysis of stochastic and nonstochastic multi-armed bandit
  problems.
\newblock \emph{Foundations and Trends{\textregistered} in Machine Learning},
  5(1):1--122.

\bibitem[{Caruana(1997)}]{caruana1997multitask}
Rich Caruana. 1997.
\newblock Multitask learning.
\newblock \emph{Machine learning}, 28(1):41--75.

\bibitem[{Caruana(1998)}]{caruana1998multitask}
Rich Caruana. 1998.
\newblock Multitask learning.
\newblock In \emph{Learning to learn}, pages 95--133. Springer.

\bibitem[{Chapelle and Li(2011)}]{chapelle2011empirical}
Olivier Chapelle and Lihong Li. 2011.
\newblock An empirical evaluation of thompson sampling.
\newblock In \emph{Advances in neural information processing systems}, pages
  2249--2257.

\bibitem[{Collobert and Weston(2008)}]{collobert2008unified}
Ronan Collobert and Jason Weston. 2008.
\newblock A unified architecture for natural language processing: Deep neural
  networks with multitask learning.
\newblock In \emph{Proceedings of the 25th international conference on Machine
  learning}, pages 160--167. ACM.

\bibitem[{Conneau et~al.(2017)Conneau, Kiela, Schwenk, Barrault, and
  Bordes}]{conneau2017supervised}
Alexis Conneau, Douwe Kiela, Holger Schwenk, Loic Barrault, and Antoine Bordes.
  2017.
\newblock Supervised learning of universal sentence representations from
  natural language inference data.
\newblock \emph{arXiv preprint arXiv:1705.02364}.

\bibitem[{Cully et~al.(2018)Cully, Chatzilygeroudis, Allocati, and
  Mouret}]{cully2018limbo}
A.~Cully, K.~Chatzilygeroudis, F.~Allocati, and J.-B. Mouret. 2018.
\newblock {Limbo: A Flexible High-performance Library for Gaussian Processes
  modeling and Data-Efficient Optimization}.
\newblock \emph{{The Journal of Open Source Software}}, 3(26):545.

\bibitem[{Dai et~al.(2016)Dai, He, and Sun}]{dai2016instance}
Jifeng Dai, Kaiming He, and Jian Sun. 2016.
\newblock Instance-aware semantic segmentation via multi-task network cascades.
\newblock In \emph{Proceedings of the IEEE Conference on Computer Vision and
  Pattern Recognition}, pages 3150--3158.

\bibitem[{Duh et~al.(2013)Duh, Neubig, Sudoh, and Tsukada}]{duh2013adaptation}
Kevin Duh, Graham Neubig, Katsuhito Sudoh, and Hajime Tsukada. 2013.
\newblock Adaptation data selection using neural language models: Experiments
  in machine translation.
\newblock In \emph{Proceedings of the 51st Annual Meeting of the Association
  for Computational Linguistics (Volume 2: Short Papers)}, volume~2, pages
  678--683.

\bibitem[{Girshick(2015)}]{girshick2015fast}
Ross Girshick. 2015.
\newblock Fast r-cnn.
\newblock In \emph{Proceedings of the IEEE international conference on computer
  vision}, pages 1440--1448.

\bibitem[{Golovin et~al.(2017)Golovin, Solnik, Moitra, Kochanski, Karro, and
  Sculley}]{golovin2017google}
Daniel Golovin, Benjamin Solnik, Subhodeep Moitra, Greg Kochanski, John Karro,
  and D~Sculley. 2017.
\newblock Google vizier: A service for black-box optimization.
\newblock In \emph{Proceedings of the 23rd ACM SIGKDD International Conference
  on Knowledge Discovery and Data Mining}, pages 1487--1495. ACM.

\bibitem[{Graves et~al.(2017)Graves, Bellemare, Menick, Munos, and
  Kavukcuoglu}]{graves2017automated}
Alex Graves, Marc~G Bellemare, Jacob Menick, Remi Munos, and Koray Kavukcuoglu.
  2017.
\newblock Automated curriculum learning for neural networks.
\newblock \emph{arXiv preprint arXiv:1704.03003}.

\bibitem[{Guo et~al.(2018)Guo, Pasunuru, and Bansal}]{guo2018dynamic}
Han Guo, Ramakanth Pasunuru, and Mohit Bansal. 2018.
\newblock Dynamic multi-level multi-task learning for sentence simplification.
\newblock \emph{arXiv preprint arXiv:1806.07304}.

\bibitem[{Hashimoto et~al.(2017)Hashimoto, Xiong, Tsuruoka, and
  Socher}]{hashimoto2017ajm}
Kazuma Hashimoto, Caiming Xiong, Yoshimasa Tsuruoka, and Richard Socher. 2017.
\newblock A joint many-task model: Growing a neural network for multiple nlp
  tasks.
\newblock In \emph{EMNLP}.

\bibitem[{Hoffman et~al.(2011)Hoffman, Brochu, and
  de~Freitas}]{hoffman2011portfolio}
Matthew~D Hoffman, Eric Brochu, and Nando de~Freitas. 2011.
\newblock Portfolio allocation for bayesian optimization.
\newblock In \emph{UAI}, pages 327--336. Citeseer.

\bibitem[{Houthooft et~al.(2016)Houthooft, Chen, Duan, Schulman, De~Turck, and
  Abbeel}]{houthooft2016vime}
Rein Houthooft, Xi~Chen, Yan Duan, John Schulman, Filip De~Turck, and Pieter
  Abbeel. 2016.
\newblock Vime: Variational information maximizing exploration.
\newblock In \emph{Advances in Neural Information Processing Systems}, pages
  1109--1117.

\bibitem[{Jaderberg et~al.(2016)Jaderberg, Mnih, Czarnecki, Schaul, Leibo,
  Silver, and Kavukcuoglu}]{jaderberg2016reinforcement}
Max Jaderberg, Volodymyr Mnih, Wojciech~Marian Czarnecki, Tom Schaul, Joel~Z
  Leibo, David Silver, and Koray Kavukcuoglu. 2016.
\newblock Reinforcement learning with unsupervised auxiliary tasks.
\newblock \emph{arXiv preprint arXiv:1611.05397}.

\bibitem[{Kaelbling et~al.(1996)Kaelbling, Littman, and
  Moore}]{kaelbling1996reinforcement}
Leslie~Pack Kaelbling, Michael~L Littman, and Andrew~W Moore. 1996.
\newblock Reinforcement learning: A survey.
\newblock \emph{Journal of artificial intelligence research}, 4:237--285.

\bibitem[{Kaiser et~al.(2017)Kaiser, Gomez, Shazeer, Vaswani, Parmar, Jones,
  and Uszkoreit}]{kaiser2017one}
Lukasz Kaiser, Aidan~N Gomez, Noam Shazeer, Ashish Vaswani, Niki Parmar, Llion
  Jones, and Jakob Uszkoreit. 2017.
\newblock One model to learn them all.
\newblock \emph{arXiv preprint arXiv:1706.05137}.

\bibitem[{Kendall et~al.(2017)Kendall, Gal, and Cipolla}]{kendall2017multi}
Alex Kendall, Yarin Gal, and Roberto Cipolla. 2017.
\newblock Multi-task learning using uncertainty to weigh losses for scene
  geometry and semantics.
\newblock \emph{arXiv preprint arXiv:1705.07115}, 3.

\bibitem[{Kingma and Ba(2014)}]{Kingma2014AdamAM}
Diederik~P. Kingma and Jimmy Ba. 2014.
\newblock Adam: A method for stochastic optimization.
\newblock \emph{CoRR}, abs/1412.6980.

\bibitem[{Knudde et~al.(2017)Knudde, {van der Herten}, Dhaene, and
  Couckuyt}]{GPflowOpt2017}
Nicolas Knudde, Joachim {van der Herten}, Tom Dhaene, and Ivo Couckuyt. 2017.
\newblock {{GP}flow{O}pt: {A} {B}ayesian {O}ptimization {L}ibrary using
  Tensor{F}low}.
\newblock \emph{arXiv preprint -- arXiv:1711.03845}.

\bibitem[{Luong et~al.(2015)Luong, Le, Sutskever, Vinyals, and
  Kaiser}]{luong2015multi}
Minh-Thang Luong, Quoc~V Le, Ilya Sutskever, Oriol Vinyals, and Lukasz Kaiser.
  2015.
\newblock Multi-task sequence to sequence learning.
\newblock \emph{arXiv preprint arXiv:1511.06114}.

\bibitem[{McCann et~al.(2018)McCann, Keskar, Xiong, and
  Socher}]{mccann2018natural}
Bryan McCann, Nitish~Shirish Keskar, Caiming Xiong, and Richard Socher. 2018.
\newblock The natural language decathlon: Multitask learning as question
  answering.
\newblock \emph{arXiv preprint arXiv:1806.08730}.

\bibitem[{Misra et~al.(2016)Misra, Shrivastava, Gupta, and
  Hebert}]{misra2016cross}
Ishan Misra, Abhinav Shrivastava, Abhinav Gupta, and Martial Hebert. 2016.
\newblock Cross-stitch networks for multi-task learning.
\newblock In \emph{Proceedings of the IEEE Conference on Computer Vision and
  Pattern Recognition}, pages 3994--4003.

\bibitem[{Moore and Lewis(2010)}]{moore2010intelligent}
Robert~C Moore and William Lewis. 2010.
\newblock Intelligent selection of language model training data.
\newblock In \emph{Proceedings of the ACL 2010 conference short papers}, pages
  220--224. Association for Computational Linguistics.

\bibitem[{Parisotto et~al.(2015)Parisotto, Ba, and
  Salakhutdinov}]{parisotto2015actor}
Emilio Parisotto, Jimmy~Lei Ba, and Ruslan Salakhutdinov. 2015.
\newblock Actor-mimic: Deep multitask and transfer reinforcement learning.
\newblock \emph{arXiv preprint arXiv:1511.06342}.

\bibitem[{Pasunuru and Bansal(2017)}]{pasunuru2017multitask}
Ramakanth Pasunuru and Mohit Bansal. 2017.
\newblock Multi-task video captioning with video and entailment generation.
\newblock \emph{arXiv preprint arXiv:1704.07489}.

\bibitem[{Pasunuru et~al.(2017)Pasunuru, Guo, and
  Bansal}]{Pasunuru2017TowardsIA}
Ramakanth Pasunuru, Han Guo, and Mohit Bansal. 2017.
\newblock Towards improving abstractive summarization via entailment
  generation.
\newblock In \emph{NFiS@EMNLP}.

\bibitem[{Peters et~al.(2018)Peters, Neumann, Iyyer, Gardner, Clark, Lee, and
  Zettlemoyer}]{peters2018Deep}
Matthew~E. Peters, Mark Neumann, Mohit Iyyer, Matt Gardner, Christopher Clark,
  Kenton Lee, and Luke Zettlemoyer. 2018.
\newblock Deep contextualized word representations.
\newblock In \emph{Proc. of NAACL}.

\bibitem[{Raj and Kalyani(2017)}]{raj2017taming}
Vishnu Raj and Sheetal Kalyani. 2017.
\newblock Taming non-stationary bandits: A bayesian approach.
\newblock \emph{arXiv preprint arXiv:1707.09727}.

\bibitem[{Rasmussen(2004)}]{rasmussen2004gaussian}
Carl~Edward Rasmussen. 2004.
\newblock Gaussian processes in machine learning.
\newblock In \emph{Advanced lectures on machine learning}, pages 63--71.
  Springer.

\bibitem[{Ruder et~al.(2017{\natexlab{a}})Ruder, Bingel, Augenstein, and
  S{\o}gaard}]{ruder2017learning}
Sebastian Ruder, Joachim Bingel, Isabelle Augenstein, and Anders S{\o}gaard.
  2017{\natexlab{a}}.
\newblock Learning what to share between loosely related tasks.
\newblock \emph{arXiv preprint arXiv:1705.08142}.

\bibitem[{Ruder et~al.(2017{\natexlab{b}})Ruder, Bingel, Augenstein, and
  S{\o}gaard}]{ruder2017sluice}
Sebastian Ruder, Joachim Bingel, Isabelle Augenstein, and Anders S{\o}gaard.
  2017{\natexlab{b}}.
\newblock Sluice networks: Learning what to share between loosely related
  tasks.
\newblock \emph{arXiv preprint arXiv:1705.08142}.

\bibitem[{Ruder and Plank(2017)}]{ruder2017learning2}
Sebastian Ruder and Barbara Plank. 2017.
\newblock Learning to select data for transfer learning with bayesian
  optimization.
\newblock \emph{arXiv preprint arXiv:1707.05246}.

\bibitem[{Russo et~al.(2018)Russo, Van~Roy, Kazerouni, Osband, Wen
  et~al.}]{russo2018tutorial}
Daniel~J Russo, Benjamin Van~Roy, Abbas Kazerouni, Ian Osband, Zheng Wen,
  et~al. 2018.
\newblock A tutorial on thompson sampling.
\newblock \emph{Foundations and Trends{\textregistered} in Machine Learning},
  11(1):1--96.

\bibitem[{Schulz et~al.(2018)Schulz, Speekenbrink, and
  Krause}]{schulz2018tutorial}
Eric Schulz, Maarten Speekenbrink, and Andreas Krause. 2018.
\newblock A tutorial on gaussian process regression: Modelling, exploring, and
  exploiting functions.
\newblock \emph{Journal of Mathematical Psychology}, 85:1--16.

\bibitem[{Shahriari et~al.(2016)Shahriari, Swersky, Wang, Adams, and
  De~Freitas}]{shahriari2016taking}
Bobak Shahriari, Kevin Swersky, Ziyu Wang, Ryan~P Adams, and Nando De~Freitas.
  2016.
\newblock Taking the human out of the loop: A review of bayesian optimization.
\newblock \emph{Proceedings of the IEEE}, 104(1):148--175.

\bibitem[{Sharma and Ravindran(2017)}]{sharma2017online}
Sahil Sharma and Balaraman Ravindran. 2017.
\newblock Online multi-task learning using active sampling.
\newblock In \emph{ICLR}.

\bibitem[{Snoek et~al.(2012)Snoek, Larochelle, and Adams}]{snoek2012practical}
Jasper Snoek, Hugo Larochelle, and Ryan~P Adams. 2012.
\newblock Practical bayesian optimization of machine learning algorithms.
\newblock In \emph{Advances in neural information processing systems}, pages
  2951--2959.

\bibitem[{S{\o}gaard(2011)}]{sogaard2011data}
Anders S{\o}gaard. 2011.
\newblock Data point selection for cross-language adaptation of dependency
  parsers.
\newblock In \emph{Proceedings of the 49th Annual Meeting of the Association
  for Computational Linguistics: Human Language Technologies: short
  papers-Volume 2}, pages 682--686. Association for Computational Linguistics.

\bibitem[{Srinivas et~al.(2009)Srinivas, Krause, Kakade, and
  Seeger}]{srinivas2009gaussian}
Niranjan Srinivas, Andreas Krause, Sham~M Kakade, and Matthias Seeger. 2009.
\newblock Gaussian process optimization in the bandit setting: No regret and
  experimental design.
\newblock \emph{arXiv preprint arXiv:0912.3995}.

\bibitem[{Swersky et~al.(2013)Swersky, Snoek, and Adams}]{swersky2013multi}
Kevin Swersky, Jasper Snoek, and Ryan~P Adams. 2013.
\newblock Multi-task bayesian optimization.
\newblock In \emph{Advances in neural information processing systems}, pages
  2004--2012.

\bibitem[{Teh et~al.(2017)Teh, Bapst, Czarnecki, Quan, Kirkpatrick, Hadsell,
  Heess, and Pascanu}]{teh2017distral}
Yee Teh, Victor Bapst, Wojciech~M Czarnecki, John Quan, James Kirkpatrick, Raia
  Hadsell, Nicolas Heess, and Razvan Pascanu. 2017.
\newblock Distral: Robust multitask reinforcement learning.
\newblock In \emph{Advances in Neural Information Processing Systems}, pages
  4496--4506.

\bibitem[{Tsvetkov et~al.(2016)Tsvetkov, Faruqui, Ling, MacWhinney, and
  Dyer}]{tsvetkov2016learning}
Yulia Tsvetkov, Manaal Faruqui, Wang Ling, Brian MacWhinney, and Chris Dyer.
  2016.
\newblock Learning the curriculum with bayesian optimization for task-specific
  word representation learning.
\newblock \emph{arXiv preprint arXiv:1605.03852}.

\bibitem[{Wang et~al.(2018)Wang, Singh, Michael, Hill, Levy, and
  Bowman}]{wang2018glue}
Alex Wang, Amapreet Singh, Julian Michael, Felix Hill, Omer Levy, and Samuel~R
  Bowman. 2018.
\newblock Glue: A multi-task benchmark and analysis platform for natural
  language understanding.
\newblock \emph{arXiv preprint arXiv:1804.07461}.

\bibitem[{van~der Wees et~al.(2017)van~der Wees, Bisazza, and
  Monz}]{van2017dynamic}
Marlies van~der Wees, Arianna Bisazza, and Christof Monz. 2017.
\newblock Dynamic data selection for neural machine translation.
\newblock \emph{arXiv preprint arXiv:1708.00712}.

\bibitem[{Xiao et~al.(2018)Xiao, Zhang, and Chen}]{xiao2018gated}
Liqiang Xiao, Honglun Zhang, and Wenqing Chen. 2018.
\newblock Gated multi-task network for text classification.
\newblock In \emph{Proceedings of the 2018 Conference of the North American
  Chapter of the Association for Computational Linguistics: Human Language
  Technologies, Volume 2 (Short Papers)}, volume~2, pages 726--731.

\end{thebibliography}
\bibliographystyle{acl_natbib}

\end{document}